\RequirePackage[svgnames]{xcolor}

\documentclass[11pt,letterpaper]{mystyle}

\usepackage[all]{hypcap}
\usepackage[comma,authoryear,compress]{natbib}
\hypersetup{
    colorlinks = true,
    citecolor = {teal},
}

\usepackage{algorithm}
\usepackage{algorithmicx}
\usepackage{algpseudocode}
\usepackage{microtype}
\usepackage{graphicx}
\usepackage{booktabs}
\usepackage{float}
\usepackage{bigstrut}

\usepackage{amsmath}
\usepackage{amssymb}
\usepackage{mathtools}
\usepackage{amsthm}
\usepackage{mathrsfs}
\usepackage{nicefrac}
\usepackage{dsfont}
\usepackage{enumitem}
\usepackage{subcaption}
\usepackage{cleveref}
\usepackage{bxcoloremoji}
\usepackage{datetime2}

\usepackage{url}
\usepackage{amsfonts}
\usepackage{fdsymbol}
\usepackage{wrapfig}
\usepackage{stackengine}
\usepackage[font=small,labelfont=bf]{caption}
\usepackage{adjustbox}
\usepackage{multirow, multicol}

\usepackage{rotating}
\usepackage{makecell}

\usepackage{tcolorbox}
\tcbuselibrary{listings,skins,breakable}
\usepackage{listings}

\newtcbinputlisting{\promptbox}[2]{%
  enhanced, breakable, listing only, listing file={#2}, title={#1},
  colback=black!3, colframe=black!65, coltitle=white, fonttitle=\bfseries\footnotesize,
  boxrule=0.5pt, arc=1.5pt, left=5pt, right=5pt, top=3pt, bottom=3pt,
  listing options={basicstyle=\ttfamily\fontsize{7pt}{8.4pt}\selectfont, columns=fullflexible, keepspaces=true,
    breaklines=true, breakindent=0pt, showstringspaces=false, upquote=true, frame=none, xleftmargin=0pt,
    aboveskip=0pt, belowskip=0pt}}

\DeclareMathOperator{\Exec}{Exec}

\newcommand{\cmark}{\ding{51}}
\newcommand{\xmark}{\ding{55}}
\newcommand{\pmark}{$\sim$}
\newcommand{\headii}[1]{\multirow{2}{*}[-0.5\dimexpr\aboverulesep+\belowrulesep+\cmidrulewidth\relax]{#1}}
\newcommand{\headiii}[1]{\multirow{3}{*}[-\dimexpr\aboverulesep+\belowrulesep+\cmidrulewidth\relax]{#1}}
\newcommand{\tdelta}[1]{{\fontsize{6.5pt}{8pt}\selectfont\textcolor{black!55}{#1}}}
\newcommand{\below}[1]{\textcolor{red!70!black}{#1}}
\newcommand{\muted}[1]{\textcolor{black!55}{#1}}

\title{\LARGE Rubrics from the Task, Verdicts from Evidence:\\
  Pointwise Rubric Rewards with Tool-Augmented Judges}

\runningtitle{Rubrics from the Task, Verdicts from Evidence: Pointwise Rubric Rewards with Tool-Augmented Judges}

\author{
Tianci Liu\textsuperscript{1} \quad
Haoyu Wang\textsuperscript{2} \\
\textsuperscript{1}University of Tennessee \quad
\textsuperscript{2}University at Albany
}

\begin{document}

\begin{abstract}
Rubric-based reward models (RMs) evaluate responses against explicit criteria. However, existing methods often derive criteria from observed response candidates, introducing response-induced bias that may limit rubric generalizability. 
Their judges also struggle with mechanical constraints when relying on reading alone, 
and commonly used pairwise judges provide only relative preferences, complicating their direct use for policy optimization. 
We propose ReVeR, a pointwise rubric RM that derives hard rules and graded principles by reasoning about the instruction alone and evaluates responses independently through tool-augmented judging. Both components are trained through label-blind distillation, in which preference labels only filter a teacher's outputs, yielding ReVeR-Traces, a large-scale dataset of rubric-reasoning and judging trajectories. Experiments demonstrate that ReVeR outperforms pairwise rubric RMs in average preference accuracy, most clearly on instruction following, as well as in test-time scaling and downstream policy optimization. We further find that generator RL improves preference accuracy while reducing rubric fidelity, and that higher preference accuracy does not necessarily translate into better downstream policies. These findings highlight that rubric fidelity, preference accuracy, and downstream utility are distinct properties, shedding light on future RM research.
\looseness=-1

\vspace{2mm}


\vspace{5mm}

\coloremojicode{1F4C5} \textbf{Date}: \today

\coloremojicode{1F917} \textbf{Model Weights \& Checkpoints}: \href{https://huggingface.co/OpenRubrics/ReVeR}{https://huggingface.co/OpenRubrics/ReVeR}

\coloremojicode{1F4DA} \textbf{Datasets}: \href{https://huggingface.co/datasets/OpenRubrics/ReVeR-Traces-Rubric}{https://huggingface.co/datasets/OpenRubrics/ReVeR-Traces-Rubric};\\
\phantom{\coloremojicode{1F4DA} \textbf{Datasets}: }\href{https://huggingface.co/datasets/OpenRubrics/ReVeR-Traces-Judge}{https://huggingface.co/datasets/OpenRubrics/ReVeR-Traces-Judge}

\coloremojicode{1F4E7} \textbf{Contact}: \href{mailto:tliu43@utk.edu}{tliu43@utk.edu}; \href{mailto:hwang28@albany.edu}{hwang28@albany.edu}

\end{abstract}
\maketitle
\vspace{3mm}
\section{Introduction}
\label{sec:intro}

Rubric-based reward models (RMs) are attracting growing attention in LLM post-training for non-verifiable tasks that lack a single correct answer. On such tasks, traditional RMs that assign holistic scalar scores~\citep{christiano2017deep,ouyang2022training,wang2024interpretable,liu2026skywork} can be unreliable~\citep{hashemi2024llm,lee2025checkeval}. Rubric-based RMs instead evaluate LLM responses against a set of explicit criteria before producing an overall evaluation~\citep{gunjal2025rubrics,viswanathan2025checklists,liu-etal-2026-openrubrics}.

Typical rubric-based RMs consist of two components: a rubric generator that proposes evaluation criteria for a given instruction, and a judge that evaluates responses against them.
To generate discriminative rubrics, existing methods often rely on observed responses.
For example, \citet{liu-etal-2026-openrubrics,xu2026alternating} first ask a teacher LLM to contrast preferred and rejected responses and propose criteria that distinguish them.
A student rubric generator is then fine-tuned on the resulting rubric dataset to predict criteria from the instruction alone.
Other approaches analyze multiple candidate responses of varying quality to identify differences or common failure modes, then construct rubrics to distinguish these responses~\citep{viswanathan2025checklists,zhang2025chasing}.
Once such a \emph{response-induced} rubric is generated, a judge evaluates candidate responses against its criteria. To improve reward-modeling performance, some methods employ pairwise judges that compare two responses side by side and determine which better satisfies the criteria~\citep{liu-etal-2026-openrubrics,xu2026alternating}.

Despite their effectiveness, existing rubric-based RMs face limitations in both rubric construction and judging.
First, response-induced rubric construction may bias the resulting criteria toward the observed candidates, potentially overlooking task requirements that all candidates satisfy or violate. This bias may limit the generalizability of the resulting rubrics to unseen responses.
Moreover, a rubric generator trained on such rubrics may inherit this bias even when it generates criteria from the instruction alone at inference time.
Second, complex instructions may involve both mechanical constraints, such as \emph{word limits}, and subjective quality dimensions, such as \emph{vividness of language}.
However, existing LLM-based judges often evaluate responses by reading them alone.
Consequently, even when a rubric correctly identifies mechanical constraints, the judge may still struggle to verify whether a response satisfies them~\citep{zeng2024evaluating,zhang2024counting}.
Furthermore, pairwise judges provide relative preferences rather than independent scores for individual responses, complicating their direct use as rewards in policy optimization.

In light of these limitations, we propose ReVeR, a pointwise rubric RM.
Instead of contrasting responses, our rubric generator derives evaluation criteria by \emph{reasoning} about the instruction alone, identifying the intended deliverable, explicit and implied requirements, and potential failure modes. These criteria are organized into \emph{hard rules} for mandatory requirements and \emph{principles} for graded quality assessment.
Moreover, ReVeR's judge evaluates each response independently under a shared, instruction-specific rubric and uses a sandboxed Python interpreter to verify mechanical constraints while relying on reasoning to assess semantic criteria.
The final reward prioritizes hard-rule satisfaction, using graded principle scores to distinguish responses satisfying the same number of hard rules.

We train ReVeR through \emph{label-blind distillation} from a stronger teacher LLM.
Specifically, the teacher first generates rubric reasoning from the instruction alone and judges the preferred and rejected responses in separate episodes.
Afterwards, the preference labels only select the examples whose scores agree with them and never alter a rubric or a judgment.
The teacher thus generates its outputs using only information available to the corresponding students at inference time.
This process yields \emph{ReVeR-Traces}, a dataset containing 30K rubric-reasoning traces and 61K tool-augmented judging trajectories.
We further train the rubric generator through a two-stage process, consisting of supervised fine-tuning (ReVeR~(SFT)) followed by reinforcement learning (ReVeR~(G-RL)).

With 4B models, ReVeR achieves higher average preference accuracy and better overall downstream policy performance under DPO and GRPO than 8B pairwise rubric RMs, most clearly on instruction following. Meanwhile, our experiments reveal that these evaluation dimensions do not necessarily align: generator RL improves preference accuracy while reducing rubric fidelity, and under GRPO the most accurate pairwise rubric RM yields a policy that asserts its own compliance and falls below the base policy on IFEval. These findings highlight that rubric fidelity, preference accuracy, and downstream utility are distinct properties that should be evaluated separately.
Our contributions are:

\begin{itemize}[leftmargin=*,itemsep=1pt,topsep=3pt]
\item We present ReVeR, a 4B pointwise rubric RM that derives typed rubrics from the instruction alone and employs a tool-augmented judge.
\item We introduce ReVeR-Traces, a large and diverse dataset containing rich rubric-reasoning traces and judging trajectories, which is released along with our model checkpoints.
\item We show that ReVeR achieves higher average preference accuracy than pairwise rubric RMs, leading on every instruction-following benchmark, and achieves stronger average downstream performance under DPO and GRPO. We also reveal that rubric fidelity, preference accuracy, and downstream utility are distinct properties, shedding light on future RM research.
\end{itemize}

\section{Related Work}
\label{sec:related}

\begin{table*}[t]
  \centering
  \caption{ReVeR and ReVeR-Traces in context.
    \textbf{(a)} \emph{From task}: criteria are written from the instruction alone;
    \emph{point.}: each response is judged on its own;
    \emph{per crit.}: one verdict per criterion;
    \emph{exec.}: verdicts use executed code.
    \textbf{(b)} \emph{Label-blind}: no preference label or competing response is in context during annotation;
    \emph{reas.}: the reasoning behind each rubric or verdict is recorded;
    \emph{tools}: programs and their outputs.
    \pmark~in part; --~not applicable.}
  \label{tab:related}
  \fontsize{8pt}{9.6pt}\selectfont
  \setlength{\tabcolsep}{3.5pt}
    \begin{tabular}[t]{@{}lccccc@{}}
      \multicolumn{6}{@{}c@{}}{\textbf{(a) Systems}} \\[2pt]
      \toprule
      & \textbf{Rubric} & \multicolumn{3}{c}{\textbf{Verdict}} & \textbf{Judge} \\
      \cmidrule(lr){2-2}\cmidrule(lr){3-5}\cmidrule(l){6-6}
      & from task & point. & per crit. & exec. & trained \\
      \midrule
      Rubric-RM\,/\,-ARM & \xmark & \xmark & \xmark & \xmark & \cmark \\
      RUBRIC-ARROW & \xmark & \cmark & \cmark & \xmark & \cmark \\
      CE-RM        & \cmark & \cmark & \xmark & \xmark & \cmark \\
      RLCF         & \xmark & \cmark & \cmark & \pmark & \xmark \\
      RewardAgent  & \pmark & \pmark & \pmark & \cmark & \xmark \\
      TIR-Judge    & \xmark & \cmark & \xmark & \cmark & \cmark \\
      \midrule
      \rowcolor{black!10}
      \textbf{ReVeR} & \cmark & \cmark & \cmark & \cmark & \cmark \\
      \bottomrule
    \end{tabular}\hfill
    \begin{tabular}[t]{@{}lrccccc@{}}
      \multicolumn{7}{@{}c@{}}{\textbf{(b) Open training data}} \\[2pt]
      \toprule
      & & \textbf{Label-} & \textbf{Rubric} & \multicolumn{3}{c}{\textbf{Judging}} \\
      \cmidrule(lr){4-4}\cmidrule(l){5-7}
      & size & \textbf{blind} & reas. & point. & reas. & tools \\
      \midrule
      OpenRubrics    & 74K  & \xmark & \xmark & \xmark & \cmark & \xmark \\
      RUBRIC-ARROW   & 119K & \xmark & \xmark & \cmark & \cmark & \xmark \\
      WildChecklists & 130K & \xmark & \xmark & \cmark & \xmark & \xmark \\
      RubricHub      & 110K & \xmark & \xmark & --     & --     & --     \\
      Feedback Coll. & 100K & \xmark & \xmark & \cmark & \cmark & \xmark \\
      R3             & 14K  & \cmark & \xmark & \pmark & \cmark & \xmark \\
      \midrule
      \rowcolor{black!10}
      \textbf{ReVeR-Traces} & 30K\,+\,61K & \cmark & \cmark & \cmark & \cmark & \cmark \\
      \bottomrule
    \end{tabular}
\end{table*}

\paragraph{Rubric and checklist rewards.}
Rubrics and checklists decompose evaluation into explicit criteria \citep{liu2023g,hashemi2024llm,kim2025biggen,lin2025wildbench,arora2025healthbench,starace2025paperbench}.
They serve as reinforcement-learning rewards (\citealp{gunjal2025rubrics}; RLCF, \citealp{viswanathan2025checklists}), as supervision for rubric generators and rubric-conditioned judges~\citep{liu-etal-2026-openrubrics}, and as components optimized jointly with the judge, pairwise in \citet{xu2026alternating}, and pointwise in \citet{jiang2026rubric}.
Some criteria are written from the instruction, either by a prompted model \citep{cook2024ticking} or by a trained reward model that gives each response one overall score~\citep{hu2026rm}.
Others are induced from observed responses, either from preferred and rejected answers \citep{liu-etal-2026-openrubrics,Xie2025AutoRubricLF,liu2026cdrrm} or from the failure modes of sampled candidates \citep{viswanathan2025checklists}.
In contrast, ReVeR derives its criteria from the instruction by reasoning about the task, distills that reasoning into a small generator, and returns one verdict per criterion for each response.

\paragraph{Reasoning and tool-augmented judges.}
LLM judges are prompted or trained to evaluate responses \citep{zheng2023judging,li2024generative,vu2024foundational}, and generative reward models reason before giving a verdict \citep{zhang2024generative,chen2026rm,whitehouse2026j1}.
Some systems add evidence retrieved by search \citep{wei2024long}, gathered by an agent \citep{zhuge2024agent}, or produced by executed code.
RewardAgent routes instruction constraints to a code-based verifier and combines its output with a preference model \citep{peng-etal-2025-agentic}, and TIR-Judge trains judges with tool-integrated reinforcement learning \citep{xu2026tirjudge}.
ReVeR lets the judge decide per criterion whether execution is needed and trains it by distillation alone (Table~\ref{tab:related}a).

\paragraph{Open data for training evaluators.}
Open resources for training rubric generators and judges differ in what they record (Table~\ref{tab:related}b).
\mbox{OpenRubrics} \citep{liu-etal-2026-openrubrics} pairs contrastively written rubrics with pairwise judge rationales, and RUBRIC-ARROW with per-criterion labels for single responses.
WildChecklists \citep{viswanathan2025checklists} holds checklists written from sampled responses, and RubricHub \citep{li2026rubrichub} holds rubrics written from a reference response.
\mbox{Feedback} Collection \citep{kim2024prometheus} and R3 \citep{anugraha2025r3} attach written feedback and distilled reasoning to scored responses.
ReVeR-Traces records the reasoning behind each rubric and the tool-augmented trajectory behind each verdict, both generated label-blind.

\section{Method}
\label{sec:method}

\begin{figure}[t]
  \centering
  \includegraphics[width=1.0\columnwidth]{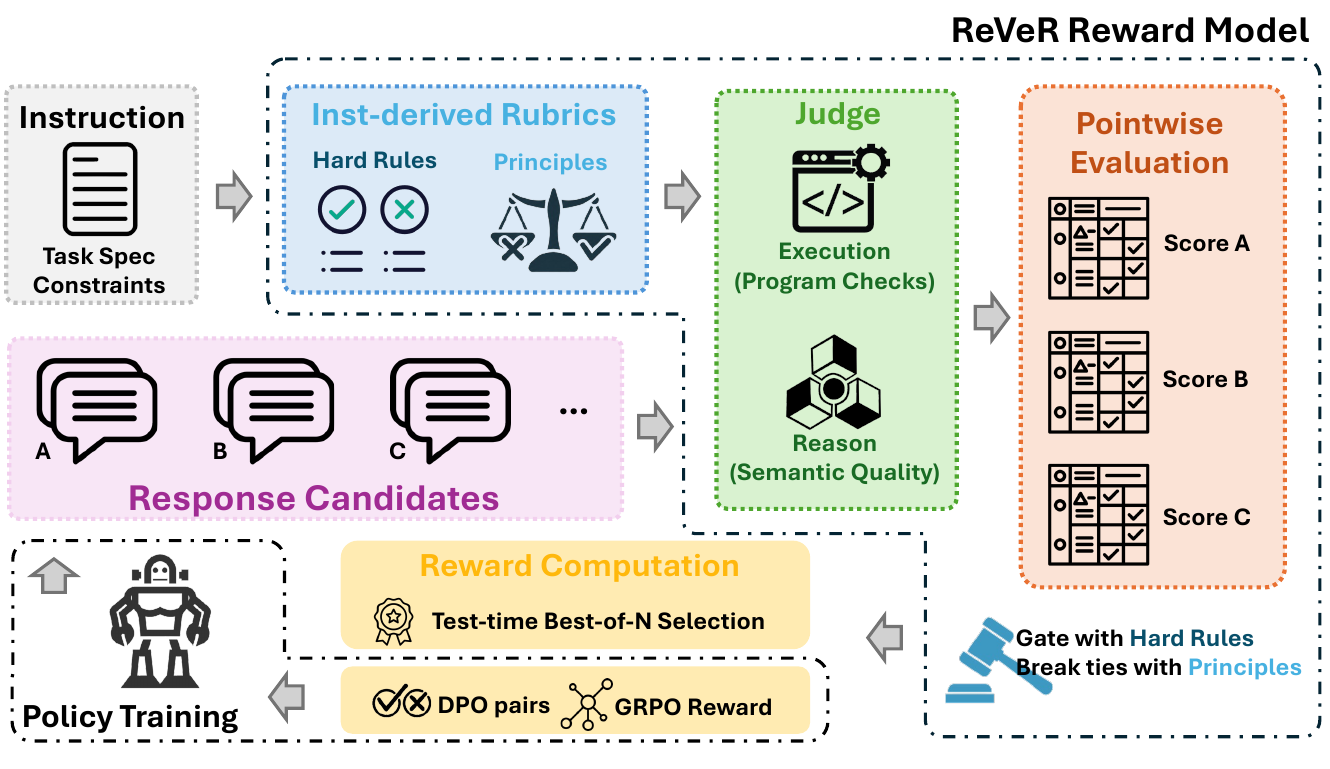}
  \caption{Overview of ReVeR.
    }
  \label{fig:overview}
\end{figure}

We propose ReVeR, which evaluates a response by rubric generation, tool-augmented verdicts, and fine-grained aggregation, as depicted in Fig.~\ref{fig:overview}.

\subsection{Problem Formulation}
\label{sec:formulation}

Let $x$ be an instruction, including any context, and $y$ a response.
A rubric $R=\{(c_i,t_i)\}_{i=1}^{k}$ is a set of natural-language criteria $c_i$, each typed as a \emph{hard rule} or a \emph{principle}.
A generator $G_\theta$ writes reasoning $z$ and a rubric; a judge $J_\phi$ with a Python interpreter produces a trajectory $\tau$ that ends in one verdict per criterion; and an aggregator $S$ maps the verdict scores $q(\tau)=(s_1,\dots,s_k)$ to a reward:
\begin{equation}
\label{eq:maps}
\begin{gathered}
  (z,R) \sim G_\theta(\,\cdot\mid x),\quad \tau \sim J_\phi(\,\cdot\mid x,y,R),\\
  r(x,y) = S\bigl(q(\tau);R\bigr).
\end{gathered}
\end{equation}
The conditioning sets in Eq.~\ref{eq:maps} determine two properties.
\emph{Prompt-level rubrics}: $R$ depends only on $x$, so one rubric is generated per prompt and shared by every response to it, and all candidates are measured against the same criteria.
\emph{Pointwise verdicts}: $\tau$ depends only on $(x,y,R)$, so $r(x,y)$ is invariant to the other responses under consideration and to the order in which they are presented, which is known to bias LLM judges \citep{wang2024large,shi2025judging}.
A pairwise judge assigns no value to a single response \citep{liusie2024llm,liu2024aligning}.

\subsection{Task-Derived Rubric Generation}
\label{sec:generator}

The generator conditions on the instruction alone and factorizes as $p_\theta(z,R\mid x)=p_\theta(z\mid x)\,p_\theta(R\mid x,z)$, reasoning about the task before writing criteria.
The reasoning $z$ identifies the deliverable the user wants, the explicit requirements and how binding each is, the expectations a competent reader infers from the task, and the ways a response could fail while appearing relevant.
Each criterion is then typed.
A hard rule is an explicit, non-negotiable requirement whose violation makes a response unacceptable; a principle is a quality dimension that a response satisfies to a degree.
The type fixes how a criterion is scored and prioritized (\S\ref{sec:judge}) and is independent of how it is checked, so a hard rule can be decided by reading.

The generator is asked to keep a criterion only if it follows from the instruction or the task type, can be met by substantively different high-quality responses, stays useful for unseen responses, captures a condition of success or failure, and is not redundant (Appendix~\ref{app:prompts}).
The tests keep the rubric a statement about the task, so it encodes functional requirements that every acceptable response meets, including those that no pair of responses separates.

\subsection{Pointwise Tool-Augmented Judging}
\label{sec:judge}

The judge scores one response per episode.
An episode is a trajectory
\begin{equation}
\label{eq:trajectory}
  \tau=\bigl((u_t,a_t,o_t)_{t=1}^{m-1},\;u_m,\;v\bigr),
\end{equation}
where $u_t$ is reasoning, $a_t$ is a Python program \citep{gao2023palprogramaidedlanguagemodels,chen2022program,yao2022react}, $o_t=\Exec(a_t)$ is its output or error, and $v$ is the final set of verdicts.
A budget $B$ bounds the number of executed programs, so $m\le B+1$.

\paragraph{Evidence.}
Whatever its type, a criterion is \emph{mechanical} if its truth is a computable function of the response string: word, sentence, or paragraph counts, prefix and case checks, forbidden terms, pattern matches, format validity, and arithmetic.
The verdict on a mechanical criterion is meant to rest on an executed output $o_t$; the verdict on a \emph{semantic} criterion, such as relevance or reasoning quality, rests on the judge's reasoning over $y$.
The judge decides which criteria need execution and consolidates their checks into one program per step, and a verdict may cite only an output that was returned.

\paragraph{Verdicts and aggregation.}
The final message $v$ gives, for every criterion, an explanation and a score: $s_i\in\{0,1\}$ for a hard rule, with insufficient evidence scored as $0$, and an integer $s_i\in\{0,\dots,100\}$ for a principle.
With $H$ and $P$ the index sets of hard rules and principles,
\begin{equation}
\label{eq:aggregate}
  S(q;R)=\sum_{i\in H}s_i+\frac{1}{101\,|P|}\sum_{j\in P}s_j .
\end{equation}
The second term is $0$ for empty $P$ and at most $\tfrac{100}{101}$, so $S$ is lexicographic.
Under the same rubric, a response that satisfies more hard rules scores higher whatever the principle scores, and principles order only responses that satisfy the same number of hard rules (Appendix~\ref{app:principle}).
We write $S(x,y;R)$ for the score of $y$ under $R$, with a subscript naming the model whose verdicts are aggregated.
Teacher trajectories use $B=1$; inference uses $B=3$, so that a failed program can be repaired (Appendix~\ref{app:training}).

\begin{figure*}[t]
  \centering
  \includegraphics[height=0.228\textwidth,trim=42bp 0 50bp 0,clip]{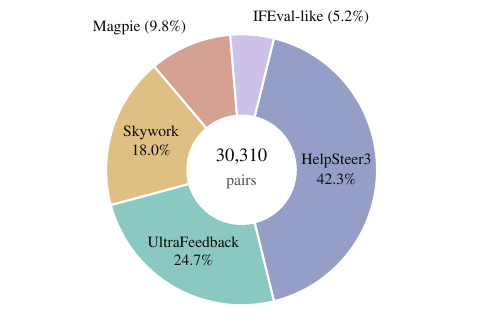}\hfill
  \includegraphics[height=0.228\textwidth]{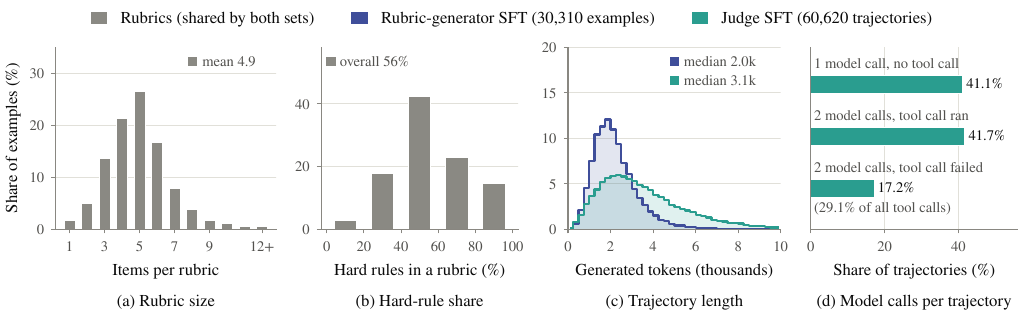}
  \caption{ReVeR-Traces.
    \textbf{Left:} sources of the 30,310 retained preference pairs.
    \textbf{Right:} (a) criteria per rubric and (b) share of hard rules, for the rubrics shared by both subsets; (c) generated tokens per rubric-reasoning trace and per judging trajectory; (d) model calls and program execution in the judging trajectories (Appendix~\ref{app:training}).}
  \label{fig:traces}
\end{figure*}

\subsection{Label-Blind Distillation}
\label{sec:distill}

Both the generator and the judge must reason, and a small model acquires that reasoning by imitating a stronger teacher $\mathcal{T}$ \citep{hinton2015distilling,hsieh2023distilling,li2023symbolic}.
A target sequence can be reproduced only from information in the student's input.
Rubric reasoning written after reading $(y^+,y^-)$ is a sample from $\mathcal{T}(\,\cdot\mid x,y^+,y^-)$, while the student models $p_\theta(z,R\mid x)$.
We therefore sample the teacher under the students' conditioning sets.
Given a preference triple $d=(x,y^+,y^-)$,
\begin{equation}
\label{eq:teacher}
  (z,R) \sim \mathcal{T}(\,\cdot\mid x),\quad \tau^{\pm} \sim \mathcal{T}(\,\cdot\mid x,y^{\pm},R),
\end{equation}
with the two responses judged in separate episodes under the same rubric.
The preference label is used once, after generation, in an acceptance test:
\begin{equation}
\label{eq:accept}
  A(d)=\mathbf{1}\bigl[\mathrm{valid}(d)\bigr]\cdot\mathbf{1}\bigl[S_{\mathcal{T}}^{+}>S_{\mathcal{T}}^{-}\bigr],
\end{equation}
where $S_{\mathcal{T}}^{\pm}=S_{\mathcal{T}}(x,y^{\pm};R)$ and $\mathrm{valid}(d)$ holds when the rubric and both trajectories are well-formed.
We call the procedure \emph{label-blind} because neither the preference label nor the competing response is in context when a rubric or a trajectory is generated.
Acceptance is rejection sampling from the teacher's own distributions.
It selects which samples are kept and leaves their content unchanged, so every retained sequence was generated only from information the student has at inference.

Each accepted triple yields one generator example $(x,z,R)$ and two judge examples $(x,y^{\pm},R,\tau^{\pm})$.
Both students are trained by maximum likelihood on the full target sequences: the generator on $(z,R)$ given $x$, and the judge on its reasoning, programs, and verdicts given $(x,y,R)$ and the preceding steps.
The judge conditions on program outputs and does not predict them.

\subsection{ReVeR-Traces}
\label{sec:traces}

ReVeR-Traces applies label-blind distillation to five preference datasets: HelpSteer3 \citep{wang2025helpsteer3preferenceopenhumanannotatedpreference}, UltraFeedback \citep{cui2024ultrafeedbackboostinglanguagemodels}, Skywork-Reward \citep{liu2024skyworkrewardbagtricksreward}, Magpie \citep{xu2024magpiealignmentdatasynthesis}, and IFEval-like instructions from OpenRubrics \citep{liu-etal-2026-openrubrics}; the teacher is DeepSeek-V4-Flash \citep{xu2026deepseek} at high reasoning effort.
Of 41,388 preference triples, 30,310 (73.2\%) pass the acceptance test, giving 30,310 rubric-reasoning traces and 60,620 judging trajectories, two per retained pair (Fig.~\ref{fig:traces}).
Each triple is sampled once, and 74.2\% of the rejections are reversed preferences, 19.5\% ties, and 6.3\% invalid outputs (Appendix~\ref{app:training}).
Rubrics average 4.9 criteria, 56\% of them hard rules, and a program is called in 58.9\% of the trajectories.
Generated label-blind, the teacher's reasoning records how it reached a judgment in place of a justification for a known outcome \citep{turpin2023language}.
We release both subsets with ReVeR (SFT), ReVeR (G-RL), and the judging harness.

\subsection{Generator Reinforcement Learning}
\label{sec:grl}

Distillation teaches the generator to write rubrics as the teacher does.
We further optimize it for the judgments those rubrics produce.
With the judge $J_\phi$ fixed, the generator is trained on preference triples to maximize
\begin{equation}
\label{eq:grl}
\begin{aligned}
  \mathcal{J}(\theta) &= \mathbb{E}_{d}\;\mathbb{E}_{(z,R)\sim G_\theta(\cdot\mid x)}\bigl[r_G(R;d)\bigr],\\
  r_G(R;d) &= \mathbf{1}\bigl[\mathrm{valid}\bigr]\cdot\mathbf{1}\bigl[S_\phi^{+}>S_\phi^{-}\bigr],
\end{aligned}
\end{equation}
where $S_\phi^{\pm}=S_\phi(x,y^{\pm};R)$ and validity requires a well-formed rubric and two valid judgments, so tied and reversed scores receive zero reward.
We optimize $\mathcal{J}$ with GRPO \citep{shao2024deepseekmath} on HelpSteer2 \citep{wang2024helpsteer2opensourcedatasettraining}, starting from the distilled generator.
The preference enters as a reward on the generator's output and never as an input, so the rubric remains a function of the instruction and the reward $r(x,y)$ remains pointwise.
Eq.~\ref{eq:grl} optimizes discrimination between two given responses, a target distinct from fidelity to the task, and \S\ref{sec:matching} measures both.
ReVeR (SFT) denotes the configuration with the distilled generator and ReVeR (G-RL) the configuration with the RL-trained generator; the two share one judge.

\section{Experiments}
\label{sec:experiments}

This section asks three questions: how accurately ReVeR ranks a preferred response above a rejected one (\S\ref{sec:rmbench}), how well its reward serves best-of-$N$ selection and policy training (\S\ref{sec:downstream}), and how closely its rubrics agree with expert rubrics (\S\ref{sec:matching}).
We compare ReVeR with the reasoning reward models RM-R1 \citep{chen2026rm}, RRM \citep{guo2026reward}, and JudgeLRM \citep{chen2025judgelrm}, the rubric-based reward models Rubric-RM, Rubric-ARM, CE-RM, and PaTaRM \citep{jian2026patarm}, and the tool-augmented TIR-Judge.
Baseline numbers are taken from the original papers where reported and computed with the released models otherwise.
Appendices~\ref{app:training} and~\ref{app:prompts} give the training and evaluation details and both prompts.

\subsection{Reward-Model Benchmarks}
\label{sec:rmbench}

\begin{table*}[t]
  \centering
  \caption{Preference accuracy (\%) on reward-model benchmarks.
    Bold and underline mark the best and second-best single-judgment results per column, reference rows excluded.
    \emph{IF pairs}: preference pairs built on IFEval and IFBench.
    \emph{Qwen3.5-4B, untrained}: the base model of ReVeR under the same protocol.
    \emph{mj@5}: each response is judged five times and the mean scores are compared.
    Avg.\ is the mean over the eleven columns; -- is not reported.}
  \label{tab:rmbench}
  \fontsize{8pt}{8.6pt}\selectfont
  \setlength{\tabcolsep}{3.4pt}
  \begin{tabular}{@{}lcccccccccccc@{}}
    \toprule
    & \multicolumn{2}{c}{\textbf{RewardBench}} & \multicolumn{2}{c}{\textbf{IF pairs}} & \textbf{RM-Bench} & \multicolumn{2}{c}{\textbf{RewardBench 2}} & \multicolumn{2}{c}{\textbf{RubricBench}} & \multicolumn{2}{c}{\textbf{JudgeBench}} & \headii{\textbf{Avg.}} \\
    \cmidrule(lr){2-3}\cmidrule(lr){4-5}\cmidrule(lr){6-6}\cmidrule(lr){7-8}\cmidrule(lr){9-10}\cmidrule(lr){11-12}
    & Chat & Hard & IFEval & IFBench & Chat & IF & Focus & Chat & IF & Know. & Reas. & \\
    \midrule
    \multicolumn{13}{@{}l}{\textit{Reasoning reward models}} \\
    \quad JudgeLRM-7B & 92.1 & 56.1 & 46.0 & 47.5 & 55.4 & 9.4 & 29.1 & 39.4 & 39.0 & 50.3 & 54.4 & 47.2 \\
    \quad RRM-7B & 77.7 & 69.5 & 51.0 & 53.2 & 59.9 & 10.0 & 60.4 & 46.1 & 41.5 & 61.4 & 54.4 & 53.2 \\
    \quad RRM-32B & \underline{94.7} & \underline{81.1} & 60.2 & 60.8 & 73.9 & 34.4 & 83.6 & 39.9 & 46.3 & \textbf{69.5} & \textbf{79.9} & 65.8 \\
    \quad RM-R1-7B (Qwen2.5-Inst) & 83.0 & 70.0 & 55.2 & 55.2 & 64.2 & 20.6 & 76.2 & 43.9 & 44.7 & 59.4 & 58.4 & 57.3 \\
    \quad RM-R1-7B (DeepSeek-Dist) & 85.3 & 67.3 & 51.0 & 56.5 & 62.2 & 13.8 & 55.4 & 46.1 & 48.0 & 58.1 & 49.7 & 53.9 \\
    \quad RM-R1-14B (Qwen2.5-Inst) & 73.5 & 79.8 & 59.0 & 60.8 & 73.2 & 23.8 & \underline{84.6} & 43.9 & 36.6 & 61.4 & 61.7 & 59.8 \\
    \quad RM-R1-14B (DeepSeek-Dist) & 90.3 & 78.9 & 61.2 & 59.0 & 71.4 & 30.6 & 79.0 & 40.6 & 44.7 & \underline{66.6} & 74.5 & 63.3 \\
    \quad RM-R1-32B (Qwen2.5-Inst) & \textbf{95.3} & 80.3 & 60.4 & 60.4 & \textbf{75.3} & 33.1 & 84.2 & 40.1 & 43.1 & 64.0 & 65.1 & 63.8 \\
    \quad RM-R1-32B (DeepSeek-Dist) & \textbf{95.3} & \textbf{83.1} & 63.2 & 58.6 & \underline{74.2} & 36.9 & 79.2 & 40.6 & 48.0 & \underline{66.6} & 75.8 & 65.6 \\
    \midrule
    \multicolumn{13}{@{}l}{\textit{Rubric-based reward models}} \\
    \quad Rubric-RM-4B & 89.6 & 72.0 & 68.2 & 60.4 & 64.8 & 36.2 & 83.8 & 49.6 & 52.8 & 55.5 & 54.4 & 62.5 \\
    \quad Rubric-RM-8B & 88.2 & 74.1 & 67.0 & 65.4 & 65.7 & 34.4 & 82.2 & 47.7 & 53.7 & 62.2 & 62.6 & 63.9 \\
    \quad Rubric-ARM-8B & 89.4 & 79.6 & 70.8 & 65.9 & 69.2 & 41.9 & \textbf{89.4} & 49.6 & 50.4 & 60.9 & 62.4 & 66.3 \\
    \quad CE-RM-4B & 82.8 & 80.5 & 61.2 & 56.9 & 69.1 & 38.8 & 84.2 & 45.6 & 46.3 & 61.7 & \underline{76.5} & 64.0 \\
    \quad PaTaRM-8B & 90.9 & 68.6 & 54.9 & 53.9 & 58.3 & 30.0 & 72.9 & 39.1 & 40.7 & 62.7 & 65.8 & 58.0 \\
    \midrule
    \multicolumn{13}{@{}l}{\textit{Tool-augmented judges, models not released}} \\
    \quad TIR-Judge-4B & 79.7 & 66.5 & 78.9 & 65.8 & 61.8 & -- & -- & -- & -- & 60.8 & 71.7 & -- \\
    \quad TIR-Judge-8B & 78.3 & 73.9 & \underline{83.0} & 68.4 & 62.3 & -- & -- & -- & -- & 64.4 & 65.5 & -- \\
    \midrule
    \multicolumn{13}{@{}l}{\textit{ReVeR protocol, 4B}} \\
    \quad Qwen3.5-4B, untrained & 75.3 & 71.4 & 70.7 & 61.9 & 51.4 & 27.7 & 63.0 & 48.8 & 51.5 & 51.3 & 53.5 & 57.0 \\
    \rowcolor{black!10}
    \quad ReVeR (SFT) & 81.0 & 76.1 & 81.4 & \underline{73.8} & 65.9 & \underline{51.5} & 75.3 & \textbf{52.7} & \underline{61.0} & 59.3 & 71.6 & \underline{68.1} \\
    \rowcolor{black!10}
    \quad ReVeR (G-RL) & 85.4 & 74.3 & \textbf{83.3} & \textbf{75.9} & 65.4 & \textbf{55.0} & 69.6 & \underline{52.3} & \textbf{62.1} & 64.4 & 72.9 & \textbf{69.1} \\
    \midrule
    \multicolumn{13}{@{}l}{\textit{References: five judgments per response, and the teacher}} \\
    \rowcolor{black!10}
    \quad ReVeR (SFT), mj@5 & 85.8 & 82.9 & 87.6 & 77.5 & 67.3 & 53.8 & 80.0 & 55.1 & 65.0 & 65.9 & 87.2 & 73.5 \\
    \rowcolor{black!10}
    \quad ReVeR (G-RL), mj@5 & 90.5 & 78.9 & 88.4 & 80.2 & 68.3 & 59.4 & 75.4 & 52.7 & 65.0 & 69.2 & 83.9 & 73.8 \\
    \quad \muted{DeepSeek-V4-Flash (teacher)} & \muted{82.5} & \muted{81.5} & \muted{82.2} & \muted{73.4} & \muted{69.2} & \muted{53.1} & \muted{74.7} & \muted{53.9} & \muted{66.7} & \muted{69.2} & \muted{81.9} & \muted{71.7} \\
    \bottomrule
  \end{tabular}
\end{table*}

On a preference pair, ReVeR scores the two responses in separate episodes under one rubric and predicts the higher-scoring one; ties and invalid judgments count as errors.
Following \citet{liu-etal-2026-openrubrics} and \citet{xu2026alternating}, Table~\ref{tab:rmbench} reports accuracy on subsets of RewardBench \citep{lambert2025rewardbench}, RM-Bench \citep{liu2025rm}, RewardBench~2 \citep{malik2026rewardbench}, RubricBench \citep{zhou2026rubricbench}, and JudgeBench \citep{tan2025judgebench}, and on IFEval and IFBench preference pairs.

\paragraph{What training on ReVeR-Traces contributes.}
Under the same judging protocol, the untrained Qwen3.5-4B \citep{qwen3.5}, as both generator and judge, averages 57.0.
ReVeR (SFT) reaches 68.1, 11.1 points higher and 3.6 below its teacher.

\paragraph{Comparison with other reward models.}
ReVeR has the two highest averages among single-judgment reward models, 69.1 with ReVeR (G-RL) and 68.1 with ReVeR (SFT), ahead of Rubric-ARM-8B (66.3) and of the 32B reasoning models RRM (65.8) and RM-R1 (65.6).
It leads on every instruction-following column and on RubricBench Chat: 83.3 on IFEval, 75.9 on IFBench, 55.0 on the instruction-following subset of RewardBench 2, where the best other model reaches 41.9, and 52.7 and 62.1 on RubricBench, against 49.6 and 53.7.
The 32B reasoning models rank first on RewardBench, RM-Bench, and JudgeBench, with 95.3 on RewardBench Chat against 85.4 for the better ReVeR configuration, and Rubric-ARM ranks first on the Focus subset of RewardBench~2.

\paragraph{Generator RL and repeated judgments.}
With the judge shared, ReVeR (G-RL) averages 1.0 point above ReVeR (SFT).
With five judgments per response, the two averages rise to 73.5 and 73.8, above the single judgment of the teacher (71.7).

\subsection{Downstream Evaluation}
\label{sec:downstream}

\paragraph{Best-of-$N$ selection and test-time scaling.}
For each prompt of IFEval \citep{zhou2023instruction} and MulDimIF \citep{ye2026muldimif}, we sample sixteen candidates from a frozen Qwen2.5-7B-Instruct \citep{Yang2024Qwen25TR}, score each once under one rubric, and select the best of $N$ \citep{snell2024scaling}.
IFEval is the mean of its four accuracies.

\begin{figure}[t]
\centering
\includegraphics[width=0.6\columnwidth]{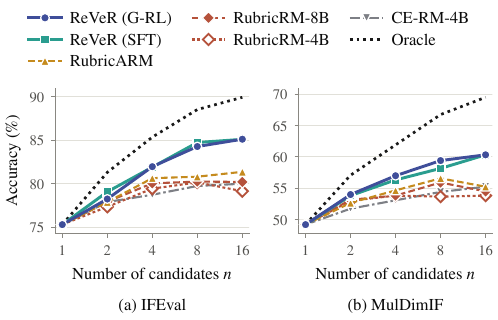}
\caption{Best-of-$N$ selection from a frozen Qwen2.5-7B-Instruct.
Each selector picks one of $n$ sampled candidates per prompt; $n=1$ is the expected score of one sampled response (no selection), and Oracle selects with the benchmark's own checker (upper bound).}
\label{fig:bon}
\end{figure}

ReVeR selects better responses than every baseline at every $N$ (Fig.~\ref{fig:bon}).
At $N=16$ both configurations reach 85.1 on IFEval and 60.3 on MulDimIF, 3.7 and 5.1 points above the strongest baseline and 9.8 and 11.1 points above a single sample.
The oracle of Fig.~\ref{fig:bon} selects with each benchmark's own checker and bounds every selector; at $N=16$ it gains 14.6 and 20.3 points over a single sample.
ReVeR obtains 67\% and 55\% of that gain, and the strongest baseline 41\% and 30\%.
With four candidates, both configurations already exceed every baseline at sixteen.

\begin{table*}[t]
\centering
\caption{Policy optimization of Qwen2.5-7B-Instruct using different RMs. Best results are in \textbf{bold}, and results below the base policy are in \below{red}.}
  \label{tab:policy}
  \fontsize{8pt}{8.8pt}\selectfont
  \setlength{\tabcolsep}{6pt}
  \begin{tabular}{@{}lccccccccl@{}}
    \toprule
    \headiii{\textbf{Reward}} & \multicolumn{4}{c}{\textbf{IFEval}} & \headiii{\textbf{IFBench}} & \headiii{\textbf{MulDimIF}} & \multicolumn{2}{c}{\textbf{Arena-Hard}} & \headiii{\textbf{Avg.}} \\
    \cmidrule(lr){2-5}\cmidrule(lr){8-9}
    & \multicolumn{2}{c}{Strict} & \multicolumn{2}{c}{Loose} & & & \headii{Style Control} & \headii{Vanilla} & \\
    \cmidrule(lr){2-3}\cmidrule(lr){4-5}
    & Prompt & Instruction & Prompt & Instruction & & & & & \\
    \midrule
    Base policy & 72.1 & 79.6 & 74.5 & 81.8 & 33.7 & 51.7 & 53.1 & 52.9 & 53.9 \\
    \midrule
    \multicolumn{10}{@{}l}{\textit{DPO}} \\
    \quad Rubric-RM-8B  & 72.6 & \below{79.5} & 78.6 & 84.8 & 35.0 & 54.3 & \below{52.5} & \below{52.2} & 55.1~\tdelta{$+$1.2} \\
    \quad Rubric-ARM-8B & 73.9 & 81.8 & 79.5 & 86.0 & 34.0 & 54.4 & 53.4 & 53.5 & 55.5~\tdelta{$+$1.6} \\
    \rowcolor{black!10}
    \quad ReVeR (SFT)  & \textbf{77.1} & \textbf{83.5} & 80.2 & 86.0 & \textbf{36.7} & \textbf{58.7} & \textbf{54.8} & 54.2 & \textbf{57.9}~\tdelta{$+$4.0} \\
    \rowcolor{black!10}
    \quad ReVeR (G-RL) & 76.7 & 83.3 & \textbf{81.9} & \textbf{87.2} & 36.4 & 56.5 & 54.0 & \textbf{54.6} & 57.4~\tdelta{$+$3.5} \\
    \midrule
    \multicolumn{10}{@{}l}{\textit{GRPO}} \\
    \quad Rubric-RM-8B  & 75.0 & 81.8 & 78.6 & 84.7 & 34.0 & \below{51.3} & 54.0 & 56.0 & 55.1~\tdelta{$+$1.2} \\
    \quad Rubric-ARM-8B & \below{63.4} & \below{73.9} & 75.4 & 82.5 & 35.4 & \below{39.3} & \textbf{55.6} & \textbf{62.0} & \below{51.8}~\tdelta{$-$2.1} \\
    \rowcolor{black!10}
    \quad ReVeR (SFT)  & \textbf{76.0} & 82.7 & \textbf{79.7} & \textbf{85.9} & 34.0 & \textbf{58.7} & \below{52.5} & \below{52.5} & 56.6~\tdelta{$+$2.7} \\
    \rowcolor{black!10}
    \quad ReVeR (G-RL) & \textbf{76.0} & \textbf{83.1} & 79.1 & 85.4 & \textbf{37.8} & 55.6 & 54.6 & 56.0 & \textbf{57.4}~\tdelta{$+$3.5} \\
    \bottomrule
  \end{tabular}
\end{table*}

\paragraph{Policy training.}
We train policies from Qwen2.5-7B-Instruct under DPO \citep{rafailov2023direct} and under GRPO, each with one of four rewards: ReVeR (SFT), ReVeR (G-RL), and the pairwise Rubric-RM-8B and Rubric-ARM-8B.
For DPO, each reward picks the best and the worst of eight sampled responses as a preference pair; for GRPO, the ReVeR score of each rollout is its reward (Appendix~\ref{app:training}).
We evaluate every policy on IFEval, IFBench \citep{pyatkin2026generalizing}, MulDimIF, and Arena-Hard v1 \citep{li2024crowdsourceddatahighqualitybenchmarks} with and without style control.

\paragraph{Preference optimization.}
Under DPO, the best value in every column of Table~\ref{tab:policy} belongs to a policy trained with ReVeR.
The average rises from 53.9 for the base policy to 57.9 with ReVeR (SFT) and 57.4 with ReVeR (G-RL), against 55.1 with Rubric-RM-8B and 55.5 with Rubric-ARM-8B.
The margin of ReVeR (SFT) over Rubric-ARM-8B is widest where compliance is checked strictly: 77.1 against 73.9 on prompt-level strict IFEval accuracy, and 58.7 against 54.4 on MulDimIF.

\paragraph{Online RL.}
Under GRPO, ReVeR again reaches the two highest averages, 57.4 with ReVeR (G-RL) and 56.6 with ReVeR (SFT), against 55.1 with Rubric-RM-8B and 51.8 with Rubric-ARM-8B.
It gives the best value in every instruction-following column.
ReVeR (G-RL) further improves Arena-Hard performance, raising scores from 52.5 with ReVeR (SFT) to 54.6 and 56.0 under the style-controlled and vanilla settings, respectively.

We also find that, under GRPO with the Rubric-ARM-8B reward, preference and compliance come apart.
Its policy reaches 62.0 on Arena-Hard without style control, 9.1 points above the base policy, and the gain shrinks to 2.5 points once style is controlled.
At the same time, its IFEval average falls to 73.8 and its MulDimIF accuracy to 39.3, below the base policy's 77.0 and 51.7, with prompt-level strict IFEval accuracy at 63.4 against 72.1.
The policy has learned to certify its own compliance, a pattern consistent with reward hacking \citep{amodei2016concrete,skalse2022defining,raina2024llm}.
Of its 541 IFEval responses, 133 end with a summary or self-assessment and 56 of these explicitly claim compliance, against 10 and 1 for the base policy, 17 and 1 with Rubric-RM-8B, and at most 23 and 2 for the policies trained with ReVeR (Appendix~\ref{app:behaviour}).
A judge without tools can check such a claim only by reading the response; ReVeR settles a word limit from the count a program returns (\S\ref{sec:judge}), which a claim of compliance cannot alter.

\subsection{Semantic Matching against Expert Rubrics}
\label{sec:matching}

\begin{table*}[t]
  \centering
  \caption{Semantic matching on RubricBench.
    \textbf{(a)} Recall (R), precision (P), and F1 (\%) of generated rubrics against expert-written rubrics.
    \textbf{(b)} Share of each criterion type in ReVeR's rubrics and the rate at which criteria of that type match an expert criterion, before and after generator RL.
    Small gray numbers are changes from ReVeR (SFT).
    On instruction following, the G-RL shares sum to 98.6\% because the output for one prompt carries no type.}
  \label{tab:matching}
  \fontsize{8pt}{8.8pt}\selectfont
  \setlength{\tabcolsep}{4.5pt}
    \begin{tabular}[t]{@{}lcccccc@{}}
      \multicolumn{7}{@{}c@{}}{\textbf{(a) Agreement with expert rubrics}} \\[2pt]
      \toprule
      & \multicolumn{3}{c}{\textbf{Chat}} & \multicolumn{3}{c}{\textbf{Instruction following}} \\
      \cmidrule(lr){2-4}\cmidrule(l){5-7}
      & R & P & F1 & R & P & F1 \\
      \midrule
      Rubric-RM-8B  & 28.9 & 20.3 & 22.8 & 43.8 & 26.4 & 32.2 \\
      Rubric-ARM-8B & 25.3 & 24.2 & 23.6 & 41.0 & 31.9 & 34.9 \\
      \midrule
      \rowcolor{black!10}
      ReVeR (SFT)  & \textbf{31.1} & \textbf{37.5} & \textbf{32.1} & \textbf{47.2} & \textbf{48.3} & \textbf{46.3} \\
      \rowcolor{black!10}
      ReVeR (G-RL) & 27.4 & 29.3 & 27.0 & 39.2 & 39.0 & 38.1 \\
      \quad\tdelta{G-RL $-$ SFT} & \tdelta{$-$3.7} & \tdelta{$-$8.2} & \tdelta{$-$5.1} & \tdelta{$-$8.0} & \tdelta{$-$9.3} & \tdelta{$-$8.2} \\
      \bottomrule
    \end{tabular}\hfill
    \begin{tabular}[t]{@{}llrr@{\hspace{3pt}}lrr@{\hspace{3pt}}l@{}}
      \multicolumn{8}{@{}c@{}}{\textbf{(b) ReVeR rubrics by criterion type}} \\[2pt]
      \toprule
      & & \multicolumn{3}{c}{\textbf{Share (\%)}} & \multicolumn{3}{c}{\textbf{Match rate (\%)}} \\
      \cmidrule(lr){3-5}\cmidrule(l){6-8}
      & & SFT & \multicolumn{2}{c}{G-RL} & SFT & \multicolumn{2}{c}{G-RL} \\
      \midrule
      Chat & Hard rule & 48.2 & 37.8 & \tdelta{$-$10.4} & 47.2 & 44.7 & \tdelta{$-$2.5} \\
           & Principle & 51.8 & 62.2 & \tdelta{$+$10.4} & 28.2 & 19.9 & \tdelta{$-$8.3} \\
      \midrule
      Instr.\ following & Hard rule & 66.1 & 52.5 & \tdelta{$-$13.6} & 60.4 & 60.2 & \tdelta{$-$0.2} \\
                        & Principle & 33.9 & 46.1 & \tdelta{$+$12.2} & 23.5 & 15.9 & \tdelta{$-$7.6} \\
      \bottomrule
    \end{tabular}
\end{table*}

RubricBench pairs each prompt with a rubric that experts derived from the instruction alone, which makes it a reference for task-derived rubrics.
On its chat and instruction-following subsets, 419 and 123 prompts, we compare the rubrics of ReVeR (SFT) and ReVeR (G-RL) with those of Rubric-RM-8B and Rubric-ARM-8B, sampling three rubrics per prompt.
The benchmark's matcher decides whether each generated criterion is semantically equivalent to an expert criterion, and we report recall, precision, and F1 against the expert rubric; no judge is run (Appendix~\ref{app:grl}).
The prompts that ReVeR's generator and judge see as in-context examples are excluded (Appendix~\ref{app:prompts}).

\paragraph{What distillation learns.}
ReVeR (SFT) is closest to the expert rubrics on all six measures (Table~\ref{tab:matching}a).
Its F1 reaches 32.1 on chat and 46.3 on instruction following, against 23.6 and 34.9 for Rubric-ARM-8B, the closer of the two baselines.
Most of the gain is in precision, 13.3 and 16.4 points above Rubric-ARM-8B, while recall is 2.2 and 3.4 points above Rubric-RM-8B, which has the higher recall of the two.
Both baselines learn their rubrics from contrasts between responses; ReVeR's distilled generator, which the pairs reach only through the acceptance test, comes closer than either to the criteria that experts write.

\paragraph{What generator RL changes.}
Once the pairs act as a reward, agreement falls on every measure, but ReVeR (G-RL) remains above both baselines in precision and F1.
Its rubrics shift from hard rules to principles (Table~\ref{tab:matching}b), with the share of hard rules falling from 48.2\% to 37.8\% on chat and from 66.1\% to 52.5\% on instruction following.
Precision falls because principles, which match expert criteria less often than hard rules, become more numerous and match less often than before, whereas hard rules match almost as often as before.
The fall in recall comes almost entirely from hard rules (Appendix~\ref{app:grl}).
On instruction following, many of the expert criteria that drop out are explicit, binary requirements such as writing entirely in Spanish, and the principles that enter name generic qualities such as accuracy.

\paragraph{Fidelity and utility.}
Lower agreement with experts leaves preference accuracy almost unchanged.
On the same RubricBench prompts, ReVeR (G-RL) has a preference accuracy of 52.3 on chat and 62.1 on instruction following, against 52.7 and 61.0 for ReVeR (SFT) (Table~\ref{tab:rmbench}).
Eq.~\ref{eq:grl} gives a rubric no reward when the two responses tie, and graded principles separate responses that binary hard rules leave tied.
RL therefore shifts rubrics towards principles, and ties fall from 3.5\% to 0.9\% of pairs on chat and from 4.1\% to 1.4\% on instruction following.
Pairs decided by principles are judged less accurately than pairs decided by hard rules, which offsets the gain from fewer ties (Appendix~\ref{app:grl}).
Generator RL optimizes discrimination between responses, and fidelity is a separate property that preference benchmarks do not measure.

\subsection{Case Study}
\label{sec:case}

\begin{figure}[t]
\centering
\fontsize{8pt}{9.6pt}\selectfont
\renewcommand{\arraystretch}{1.0}
\begin{tabular}{@{}>{\raggedright\arraybackslash}p{\columnwidth}@{}}
\toprule
\textbf{Instruction.} \ldots{} Which college would you recommend? \ldots{} be sure the letter p appears at least 15 times in your response. \\
\midrule
\textbf{Responses.} Both ask for the user's preferences and name no college. A contains 16 p's and B contains 13. \textbf{Expert label: A.} \\
\midrule
\textbf{Shared hard rule.} ReVeR: ``The response contains at least 15 occurrences of the lowercase letter `p'.'' Rubric-ARM: ``The response must contain at least 15 instances of the letter p.'' \\
\midrule
\rowcolor{black!10}
\textbf{ReVeR judge}, one program per response: \texttt{response.count('p')} returns 16 for A and 13 for B. Scores 3.10 and 2.00; it prefers \textbf{A} in 6 of 6 runs. \\
\midrule
\textbf{Rubric-ARM judge}, reading both responses. Order (A,\,B): ``only Response B satisfies the crucial letter-p count gatekeeper.'' Order (B,\,A): both ``fail \ldots{} the letter-p count gatekeeper'' and B ``at least reaches closer to 15 p's.'' It prefers \textbf{B} in both orders of 3 of 3 repeats. \\
\bottomrule
\end{tabular}
\caption{Case study on the tool-augmented judge.
Given the same criterion, ReVeR's judge succeeds by running a program and Rubric-ARM's judge fails.}
\label{fig:case}
\end{figure}

As an illustration, on the RubricBench prompt of Fig.~\ref{fig:case}, ReVeR and Rubric-ARM write the same hard rule and prefer opposite responses.
ReVeR's judge counts with a program and prefers Response A, which satisfies the rule, in all six runs; Rubric-ARM's pairwise judge reads and prefers Response B in both presentation orders of all three repeats (Appendix~\ref{app:cases}).

\section{Conclusion}
\label{sec:conclusion}

We present ReVeR, a pointwise rubric RM, and ReVeR-Traces, a collection of reasoning traces for rubric generation and tool-augmented judgment trajectories.
Its generator derives criteria by reasoning about the task, its judge reaches each verdict from the response and from evidence it executes, and both are trained by label-blind distillation.
With 4B models, ReVeR has the highest average preference accuracy on reward-model benchmarks among single-judgment reward models, and it improves on pairwise rubric rewards in agreement with expert rubrics, best-of-$N$ selection, and policy optimization.
We also find that rubric fidelity, preference accuracy, and downstream utility are distinct properties of a reward and have to be measured separately.

\section*{Acknowledgements}
This research is supported by the National Artificial Intelligence Research Resource (NAIRR) Pilot (grant NAIRR260072) through the CloudBank project.

\clearpage
\bibliography{custom}

\appendix
\section{Distribution of Principle Scores}
\label{app:principle}

\begin{figure}[t]
  \centering
  \includegraphics[width=0.6\columnwidth]{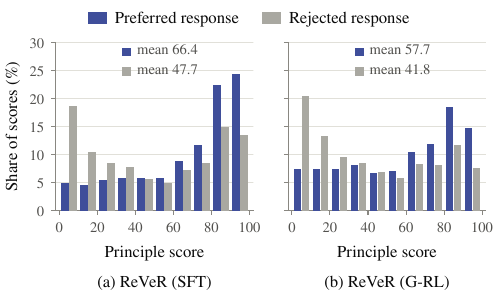}
  \caption{Distribution of the principle scores assigned by the judge, pooled over the reward-model benchmarks of Table~\ref{tab:rmbench}, with rubrics from (a) ReVeR (SFT) and (b) ReVeR (G-RL).
    Scores are integers from 0 to 100 and are shown in bins of 10; the last bin includes 100.
    Preferred and rejected follow the ground-truth preference of each benchmark.
    Each group is normalized separately, so its bars sum to 100\%.}
  \label{fig:principle}
\end{figure}

Fig.~\ref{fig:principle} shows the scores that the judge assigns to principles, pooled over the reward-model benchmarks of Table~\ref{tab:rmbench}: 17,823 scores under ReVeR (SFT) rubrics and 24,620 under ReVeR (G-RL) rubrics.

\paragraph{The judge uses the whole scale.}
The scores take more than 90 distinct values, and 1.5\% (SFT) and 0.5\% (G-RL) of them equal 100.

\paragraph{The scores carry preference signal.}
Responses that the benchmark prefers concentrate at the high end, with 47\% (SFT) and 33\% (G-RL) of their scores at 80 or above.
Rejected responses have a second mode at the bottom: 19\% and 21\% of their scores are below 10, against 5\% and 7\% for preferred responses.
The mean gap between the two groups is 18.7 and 15.9 points.

\paragraph{Principles break ties.}
The two distributions overlap: 28\% and 19\% of the scores given to rejected responses are 80 or above.
On the 41--42\% of pairs whose responses tie on hard rules, the principle average favours the preferred response in 58--59\%.
This supports the aggregation of Eq.~\ref{eq:aggregate}, which ranks responses by hard rules first and uses the principle average to break ties.
On RubricBench, pairs decided by hard rules are also judged more accurately than pairs decided by principles (Appendix~\ref{app:grl}).

\paragraph{Generator RL.}
G-RL rubrics contain more principles than SFT rubrics, 2.7 against 1.9 per response, and their principles receive lower scores, with a mean of 48.5 against 55.5.

\section{Training and Evaluation Details}
\label{app:training}

Table~\ref{tab:app-training} lists the settings for distillation (\S\ref{sec:distill}), generator RL (\S\ref{sec:grl}), and policy training (\S\ref{sec:downstream}).
Generator RL samples rubrics with thinking enabled and uses the dynamic sampling of DAPO \citep{yu2026dapo}.
For the policy, DPO uses the sigmoid loss and adds the negative log-likelihood of the preferred response with the weight given in the table, and GRPO uses the DAPO loss and excludes overlong rollouts from the loss.
Each model is trained once and evaluated multiple times, and we report the mean.

\begin{table*}[t]
  \centering
  \caption{Training settings.
    A batch counts packed sequences in distillation, pairs in DPO, and rollouts in RL.
    The length limit applies to the packed sequence in distillation and to the generated tokens in RL.
    Distillation uses a cosine schedule with 5\% warm-up; the other runs use a constant learning rate.
    The two counts of DPO pairs are for ReVeR (SFT) and ReVeR (G-RL), and a clip range $a$\,/\,$b$ gives the lower and the upper bound.}
  \label{tab:app-training}
  \fontsize{8pt}{9.6pt}\selectfont
  \setlength{\tabcolsep}{9pt}
  \newcommand{\na}{\muted{--}}
  \begin{tabular}{@{}lccccc@{}}
    \toprule
    & \multicolumn{2}{c}{\textbf{Distillation}} & \textbf{Generator RL} & \multicolumn{2}{c}{\textbf{Policy}} \\
    \cmidrule(lr){2-3}\cmidrule(lr){4-4}\cmidrule(l){5-6}
    & Generator & Judge & GRPO & DPO & GRPO \\
    \midrule
    Initial model         & \multicolumn{2}{c}{Qwen3.5-4B} & distilled generator & \multicolumn{2}{c}{Qwen2.5-7B-Instruct} \\
    Training examples     & 30,310 & 60,620 & 7,118 & 20,407\,/\,20,479 & 21,627 \\
    Epochs                & 1 & 1 & 1 & 1 & 1 \\
    \midrule
    Learning rate         & $8\times10^{-6}$ & $8\times10^{-6}$ & $1\times10^{-6}$ & $8\times10^{-7}$ & $1\times10^{-6}$ \\
    GPUs                  & 8 & 8 & 5 & 4 & 5 \\
    Per-GPU batch         & 4 & 4 & 16 & 4 & 8 \\
    Accumulation steps    & 2 & 2 & 2 & 4 & 6 \\
    Batch per step        & 64 & 64 & 160 & 64 & 240 \\
    Length limit (tokens) & 16,384 & 16,384 & 4,096 & 8,192 & 2,048 \\
    \midrule
    Prompts per step      & \na & \na & 32 & \na & 40 \\
    Rollouts per prompt   & \na & \na & 5 & \na & 6 \\
    Temperature           & \na & \na & 1.0 & \na & 1.0 \\
    Top-$p$               & \na & \na & 0.95 & \na & 0.95 \\
    Top-$k$               & \na & \na & 20 & \na & \na \\
    Presence penalty      & \na & \na & 1.5 & \na & \na \\
    \midrule
    Clip range            & \na & \na & 0.2 & \na & 0.2\,/\,0.28 \\
    KL weight             & \na & \na & $1\times10^{-3}$ & \na & $1\times10^{-4}$ \\
    DPO $\beta$           & \na & \na & \na & 0.1 & \na \\
    NLL weight            & \na & \na & \na & 0.2 & \na \\
    \bottomrule
  \end{tabular}
\end{table*}

\paragraph{Judging protocol.}
A response is scored with one rubric per prompt and one judging episode with $B=3$.
A program runs with a timeout of three seconds.
A call that fails can be repaired with a remaining call, which can also run a follow-up check.
Once the budget is spent, the judge is instructed to complete its verdicts from the response, to name the missing result in the explanations it affects, and never to claim a verification; a required property without sufficient evidence counts as not satisfied (Appendix~\ref{app:prompts}).
A trajectory whose verdicts fail validation, which requires exactly one in-range score per criterion, is resampled, and a response left without valid verdicts receives no score.
A response without a valid judgment counts as an error on preference benchmarks, is skipped in selection, and receives no reward in policy training.
The benchmark subsets of Table~\ref{tab:rmbench} follow the evaluations of Rubric-RM and Rubric-ARM \citep{liu-etal-2026-openrubrics,xu2026alternating}.

\paragraph{ReVeR-Traces records.}
A rubric record pairs an instruction with the teacher's reasoning and typed rubric; a judging record adds one response, the complete trajectory with programs and their outputs, and the per-criterion explanations and scores.
The median trace and trajectory run to 2.0K and 3.1K generated tokens (Fig.~\ref{fig:traces}).
A program fails, by an exception or a timeout, in 17.2\% of the trajectories, which is 29.1\% of the program calls.
With $B=1$ the teacher cannot repair it, so these verdicts follow the rule of the judging protocol, and the triple is retained only if its scores still reproduce the preference.
The IFEval-like instructions are part of OpenRubrics \citep{liu-etal-2026-openrubrics}, make up 1,570 of the 30,310 retained triples, and share no prompt with the IFEval and IFBench prompts used in evaluation.

\paragraph{Acceptance test.}
The acceptance test of Eq.~\ref{eq:accept} is applied once to each triple.
The teacher writes one rubric and two trajectories, and nothing is resampled when the scores disagree with the label.
Of the 11,078 rejected triples, 74.2\% reverse the preference, 19.5\% tie, and 6.3\% are invalid, so the test mostly removes teacher judgments that contradict the human label.
Ties, in which the rubric and its verdicts do not separate the two responses, are 5.2\% of all triples.
The test retains 67.9\% of the triples of HelpSteer3, 83.1\% of UltraFeedback, 75.0\% of Skywork-Reward, 71.2\% of Magpie, and 77.2\% of the IFEval-like instructions.

\paragraph{Baselines.}
We compare with three groups of systems.
\emph{Reasoning reward models} reason before giving a verdict: RM-R1, RRM, and JudgeLRM.
\emph{Rubric-based reward models} condition the verdict on generated criteria: Rubric-RM, Rubric-ARM, CE-RM, and PaTaRM.
\emph{Tool-augmented judges} are represented by TIR-Judge.
Two references show what training contributes: the untrained Qwen3.5-4B under the same judging protocol, and the teacher, DeepSeek-V4-Flash.

\paragraph{Best-of-$N$ selection.}
We generate one rubric $R$ for $x$, score each of $N$ candidates $y_1,\dots,y_N$ from a frozen policy once, in its own episode, and return the candidate with the highest valid score, $\hat y=\arg\max_{n} S_\phi(x,y_n;R)$.
Selection costs one generator call and $N$ judge episodes, each with at most three program calls of three seconds.

\paragraph{Policy training data.}
We train the policy, Qwen2.5-7B-Instruct, on HIR-16K \citep{zhang2025replay} with 4,998 open-ended chat prompts sampled from UltraChat \citep{ding2023enhancingchatlanguagemodels}, UltraFeedback \citep{cui2024ultrafeedbackboostinglanguagemodels}, and WildChat \citep{zhao2024wildchat1mchatgptinteraction}.

\paragraph{Preference optimization.}
To train a policy with DPO, we build preference pairs from the policy's own samples.
For each instruction we draw $N=8$ candidates, score each twice, and average the normalized scores $\bar S=S/(|H|+\tfrac{100}{101})\in[0,1]$.
The highest-scoring complete candidate becomes $y^+$ and the lowest-scoring candidate $y^-$, and the pair is kept when $\bar S(y^+)-\bar S(y^-)>\delta$, with $\delta=0.05$.
For the pairwise baselines, a bubble pass of pairwise comparisons selects the best and the worst candidate.

\paragraph{Online reinforcement learning.}
For GRPO, the rubric of each training prompt is generated once in advance and the judge scores every rollout online; $S_\phi(x,y;R)$ is the rollout's reward.
Every rollout of a prompt is scored against the same rubric, so the group-relative advantage compares rollouts on a common scale.
The pairwise baselines use the online reward of their original work \citep{liu-etal-2026-openrubrics,xu2026alternating}.

\section{Behaviour of the Trained Policies}
\label{app:behaviour}

Table~\ref{tab:app-behaviour} counts surface behaviours in the responses of each policy of \S\ref{sec:downstream} on IFEval (541 responses) and MulDimIF (1,200 responses).

\begin{table}[t]
  \centering
  \caption{Surface behaviours of the trained policies.
    IFEval, of 541 responses: those that end with a summary or self-assessment (closing), those among them that explicitly claim compliance (claims), and those whose first line is bold or a heading (heading).
    MulDimIF, of 1,200 responses: those that contain a compliance statement (claims), and the failure rate on capitalization constraints (caps).}
  \label{tab:app-behaviour}
  \fontsize{8pt}{9.6pt}\selectfont
  \setlength{\tabcolsep}{4pt}
  \begin{tabular}{@{}lrrrrr@{}}
    \toprule
    & \multicolumn{3}{c}{\textbf{IFEval}} & \multicolumn{2}{c}{\textbf{MulDimIF}} \\
    \cmidrule(lr){2-4}\cmidrule(l){5-6}
    & closing & claims & heading & claims & caps (\%) \\
    \midrule
    Base policy & 10 & 1 & 17 & 14 & 4 \\
    \midrule
    \multicolumn{6}{@{}l}{\textit{DPO}} \\
    \rowcolor{black!10}
    \quad ReVeR (SFT)  & 15 & 2 & 20 & 17 & 10 \\
    \rowcolor{black!10}
    \quad ReVeR (G-RL) & 16 & 0 & 21 & 24 & 7 \\
    \midrule
    \multicolumn{6}{@{}l}{\textit{GRPO}} \\
    \quad Rubric-RM  & 17  & 1  & 58  & 47  & 34 \\
    \quad Rubric-ARM & 133 & 56 & 117 & 228 & 63 \\
    \rowcolor{black!10}
    \quad ReVeR (SFT)  & 15 & 2 & 39 & 20 & 7 \\
    \rowcolor{black!10}
    \quad ReVeR (G-RL) & 23 & 2 & 43 & 25 & 4 \\
    \bottomrule
  \end{tabular}
\end{table}

Policies trained with ReVeR end between 15 and 23 responses with a summary or self-assessment and make at most 2 explicit claims of compliance, against 10 and 1 for the base policy.
The policy trained with the Rubric-ARM reward under GRPO ends 133 responses this way, 56 of them claiming compliance, opens 117 with a bold line or a heading, states compliance in 228 MulDimIF responses, and fails 63\% of the capitalization constraints.
It is the only policy that shows the pattern at this scale.
The loose IFEval criterion, which also tests a response without its first line, without its last line, and without markdown emphasis, restores 12.0 points of its prompt-level accuracy, which locates the failures in what the policy adds around its answer.
The policy trained with the Rubric-RM reward under GRPO rarely certifies itself, with 17 such endings and 1 explicit claim, and fails 34\% of the capitalization constraints.

\section{Further Analysis of Generator RL}
\label{app:grl}

This appendix details how generator RL changes ReVeR's rubrics (\S\ref{sec:matching}).
All statistics are computed on the chat (419 prompts, 2,504 expert criteria) and instruction-following (IF; 123 prompts, 702 expert criteria) subsets of RubricBench and average three sampled rubrics per prompt.
SFT and G-RL denote the generators of ReVeR (SFT) and ReVeR (G-RL), and match rates pool over criteria.
In Table~\ref{tab:matching}a, recall is the share of expert criteria matched by at least one generated criterion and precision is the share of generated criteria that match an expert criterion; both are averaged over prompts and over the three rubrics per prompt.

\paragraph{Shape of the rubrics.}
Table~\ref{tab:app-shape} shows that RL leaves the number of criteria nearly unchanged; a criterion has about 21 words before and after.
Almost every rubric now contains a principle, and the reasoning that precedes the rubric is less than half as long.
These statistics, like the type shares of Table~\ref{tab:matching}b, do not depend on the matcher.

\begin{table}[t]
  \centering
  \caption{Shape of ReVeR's rubrics before and after generator RL: criteria per rubric, share of rubrics without a principle, and length of the reasoning that precedes the rubric.}
  \label{tab:app-shape}
  \small
  \setlength{\tabcolsep}{4pt}
  \begin{tabular}{@{}lrrrr@{}}
    \toprule
    & \multicolumn{2}{c}{\textbf{Chat}} & \multicolumn{2}{c}{\textbf{IF}} \\
    \cmidrule(lr){2-3}\cmidrule(l){4-5}
    & SFT & G-RL & SFT & G-RL \\
    \midrule
    Criteria per rubric         & 4.65  & 5.00  & 5.60   & 5.47  \\
    Without a principle (\%)    & 3.9   & 0.2   & --     & --    \\
    Reasoning (characters)      & 7,826 & 3,571 & 10,190 & 4,332 \\
    \bottomrule
  \end{tabular}
\end{table}

\paragraph{Decomposition of the precision drop.}
Pooled over criteria, precision falls by 8.0 points on chat and 8.9 on instruction following.
Table~\ref{tab:app-decomp} splits each drop into the shift in type shares and the change in each type's match rate.
On chat the drop comes mainly from principles that match expert criteria less often; on instruction following it comes mainly from the shift towards principles.
The match rate of hard rules contributes less than one point on either subset.

\begin{table}[t]
  \centering
  \caption{Change in pooled precision from SFT to G-RL and its decomposition into the shift in type shares and the change in the match rate of each type.}
  \label{tab:app-decomp}
  \small
  \setlength{\tabcolsep}{4pt}
  \begin{tabular}{@{}lrrrr@{}}
    \toprule
    & & \multicolumn{3}{c}{\textbf{Contribution of}} \\
    \cmidrule(l){3-5}
    & $\Delta$\,P & shares & hard rules & principles \\
    \midrule
    Chat & $-$8.0 & $-$2.0 & $-$0.9 & $-$5.1 \\
    IF   & $-$8.9 & $-$5.3 & $-$0.1 & $-$3.5 \\
    \bottomrule
  \end{tabular}
\end{table}

\paragraph{Position within the rubric.}
A rubric lists its hard rules first and its principles after them.
On chat, the first two criteria match expert criteria about as often after RL as before, and the gap opens from the third criterion on (Table~\ref{tab:app-position}).
After RL the second criterion is a principle in 53\% of rubrics, up from 36\%.
Hard rules in late positions also weaken: at the fourth position their match rate falls from 47.7\% to 34.3\%.

\begin{table}[t]
  \centering
  \caption{Match rate (\%) of the criterion at each position of the rubric, chat subset.}
  \label{tab:app-position}
  \small
  \setlength{\tabcolsep}{6pt}
  \begin{tabular}{@{}lrrrr@{}}
    \toprule
    \textbf{Position} & 1 & 2 & 4 & 5 \\
    \midrule
    SFT  & 46.9 & 43.6 & 33.2 & 28.3 \\
    G-RL & 45.3 & 40.0 & 18.8 & 15.3 \\
    \bottomrule
  \end{tabular}
\end{table}

\paragraph{Coverage of expert criteria.}
The drop in recall comes almost entirely from hard rules.
A rubric matches 0.21 fewer expert criteria with its hard rules on chat and 0.48 fewer on instruction following, and 0.06 and 0.04 fewer with its principles; the additional principles do not restore the coverage.
Table~\ref{tab:app-coverage} reports coverage per prompt and the expert criteria that are stably lost or gained.
An expert criterion is stably lost if SFT covers it in at least two of the three samples and G-RL in none, and stably gained in the reverse case.

\begin{table}[t]
  \centering
  \caption{Coverage of expert criteria before and after generator RL: expert criteria covered per prompt, of about six, prompts on which no criterion matches, and expert criteria stably lost or gained.}
  \label{tab:app-coverage}
  \small
  \setlength{\tabcolsep}{4pt}
  \begin{tabular}{@{}lrrrr@{}}
    \toprule
    & \multicolumn{2}{c}{\textbf{Chat}} & \multicolumn{2}{c}{\textbf{IF}} \\
    \cmidrule(lr){2-3}\cmidrule(l){4-5}
    & SFT & G-RL & SFT & G-RL \\
    \midrule
    Covered per prompt          & 1.9  & 1.6  & 2.7 & 2.2 \\
    Prompts with no match (\%)  & 19.3 & 23.8 & 4.9 & 7.3 \\
    \midrule
    Stably lost under G-RL   & \multicolumn{2}{c}{146} & \multicolumn{2}{c}{53} \\
    Stably gained under G-RL & \multicolumn{2}{c}{81}  & \multicolumn{2}{c}{17} \\
    \bottomrule
  \end{tabular}
\end{table}

On instruction following, many of the lost criteria are explicit requirements of the instruction that can be decided as true or false:
\begin{itemize}[leftmargin=*,itemsep=0pt,topsep=2pt]
  \item \emph{Is the response written entirely in Spanish?}
  \item \emph{Is the response written entirely in capital letters without any lowercase or mixed-case words?}
  \item \emph{Is the final summary preceded by exactly six asterisk symbols (******) as a separator?}
  \item \emph{Does the response use the 5-line AABBA format of a limerick?}
\end{itemize}
On chat, the lost criteria concern answering the question directly and the structure of the answer:
\begin{itemize}[leftmargin=*,itemsep=0pt,topsep=2pt]
  \item \emph{Does the response directly address the user's intended question?}
  \item \emph{Does the response provide a clear step-by-step structure?}
  \item \emph{Does the response avoid inventing extensive, detailed biographical facts not implied by the input?}
\end{itemize}
The gained criteria ask for explanation:
\begin{itemize}[leftmargin=*,itemsep=0pt,topsep=2pt]
  \item \emph{Does the response provide explanations or reasons that clarify `why'?}
  \item \emph{Does the response ask clarifying questions to understand the user's intent, rather than providing only definitive interpretations?}
\end{itemize}

\paragraph{Generic criteria.}
The criteria that RL adds describe general qualities, and the ones it removes are constraints tied to the prompt.
Table~\ref{tab:app-words} lists the words whose use changes most on chat.
Criteria that carry an example, marked by \emph{e.g.}\ or \emph{such as}, rise from 39\% to 62\% of principles and from 18\% to 34\% of hard rules; a hard rule with an example matches an expert criterion 37.5\% of the time, against 48.4\% without one.
The share of a criterion's content words that also occur in the instruction falls from 27.6\% to 24.1\% for hard rules and from 10.0\% to 7.7\% for principles.
By topic, the match rate of criteria about tone, style, and clarity falls from 33.7\% to 19.9\% and that of criteria about relevance from 32.7\% to 19.8\%, while criteria about length and quantity move from 42.3\% to 38.9\%.

\begin{table}[t]
  \centering
  \caption{Share of criteria (\%) that contain each word, chat subset.}
  \label{tab:app-words}
  \small
  \setlength{\tabcolsep}{4pt}
  \begin{tabular}{@{}lrr@{\hspace{14pt}}lrr@{}}
    \toprule
    \multicolumn{3}{@{}c}{\textbf{More frequent after RL}} & \multicolumn{3}{c@{}}{\textbf{Less frequent after RL}} \\
    \cmidrule(r){1-3}\cmidrule(l){4-6}
    & SFT & G-RL & & SFT & G-RL \\
    \midrule
    accurately & 1.9 & 7.2 & any        & 9.6 & 3.8 \\
    practical  & 1.4 & 4.3 & each       & 7.1 & 4.3 \\
    organized  & 1.2 & 3.9 & explicitly & 4.8 & 1.6 \\
    actionable & 1.4 & 3.8 & directly   & 4.8 & 1.7 \\
    coherently & 0.4 & 2.7 & every      & 1.2 & 0.2 \\
    engaging   & 0.2 & 2.2 &            &     &     \\
    \bottomrule
  \end{tabular}
\end{table}

\paragraph{Preference accuracy.}
Table~\ref{tab:app-accuracy} gives the preference accuracy of the two configurations on the same prompts and how pairs are decided.
The reward of generator RL (Eq.~\ref{eq:grl}) contains no term for agreement with expert rubrics: a rubric earns 1 when the preferred response scores strictly higher and 0 on a tie or a reversal.
Coverage and accuracy are related on instruction following, where ReVeR (G-RL) is correct on 37.0\% of the pairs whose rubric covers no expert criterion, 63.6\% with two covered, and 72.0\% with three; on chat, accuracy stays between 45\% and 58\% at every level of coverage.
On the 100 chat prompts where the F1 of G-RL is at least 15 points below that of SFT, accuracy is 60.7 for ReVeR (G-RL) and 57.7 for ReVeR (SFT); on the 39 such instruction-following prompts it is 55.6 and 59.8.

\begin{table}[t]
  \centering
  \caption{Preference accuracy on RubricBench and the criteria that decide each pair. A tie counts as an error.}
  \label{tab:app-accuracy}
  \small
  \setlength{\tabcolsep}{4pt}
  \begin{tabular}{@{}lrrrr@{}}
    \toprule
    & \multicolumn{2}{c}{\textbf{Chat}} & \multicolumn{2}{c}{\textbf{IF}} \\
    \cmidrule(lr){2-3}\cmidrule(l){4-5}
    & SFT & G-RL & SFT & G-RL \\
    \midrule
    Preference accuracy               & 52.7 & 52.3 & 61.0 & 62.1 \\
    Ties (\%)                          & 3.5  & 0.9  & 4.1  & 1.4  \\
    \midrule
    Decided by hard rules (\%)         & 39.9 & 38.2 & 60.4 & 54.5 \\
    Decided by principles (\%)         & 56.3 & 60.6 & 35.0 & 43.1 \\
    \midrule
    Accuracy, hard rules decide        & 59.1 & 58.5 & 75.8 & 77.1 \\
    Accuracy, principles decide        & 51.8 & 49.5 & 43.4 & 46.5 \\
    \bottomrule
  \end{tabular}
\end{table}

\paragraph{An example.}
Table~\ref{tab:app-example} compares the rubrics that the two generators write for the chat prompt \emph{how to drink water}.
The criteria of G-RL stay on topic; they are more detailed and carry parenthetical examples, and the matcher does not judge them equivalent to the expert criteria.

\begin{table*}[t]
  \centering
  \caption{Expert criteria for the prompt \emph{how to drink water} and the closest criterion in one sampled rubric from each generator. \cmark~marks a match. SFT matches three expert criteria and G-RL one.}
  \label{tab:app-example}
  \small
  \setlength{\tabcolsep}{5pt}
  \begin{tabular}{@{}p{0.23\textwidth}cp{0.31\textwidth}cp{0.31\textwidth}@{}}
    \toprule
    \textbf{Expert criterion} & \multicolumn{2}{c}{\textbf{SFT} (2 hard rules, 3 principles)} & \multicolumn{2}{c}{\textbf{G-RL} (1 hard rule, 5 principles)} \\
    \midrule
    Has a clear step-by-step structure
      & \cmark & ``presented in an organized, coherent manner that makes it easy \ldots\ to follow''
      & \xmark & ``sufficiently clear and organized to be followed without prior specialized knowledge'' \\
    Gives appropriate health and safety cautions
      & \cmark & ``avoids recommendations that could reasonably cause harm''
      & \cmark & ``accounts for potential risks such as choking, aspiration, or dehydration'' \\
    Stays on the question, without unrelated content
      & \cmark & ``directly addresses the question \ldots\ rather than providing unrelated information''
      & \xmark & ``provides a practical method or technique \ldots\ rather than an unrelated topic'' \\
    Includes specific practical details
      & \xmark & --
      & \xmark & ``includes actionable steps (e.g., posture, swallowing rhythm, breath control)'' \\
    Reads the question as about healthy drinking habits, not the literal act
      & \xmark & --
      & \xmark & ``appropriately contextualizes the technique (e.g., casual hydration, emergency survival \ldots)'' \\
    \bottomrule
  \end{tabular}
\end{table*}

\section{Case Studies}
\label{app:cases}

This appendix gives two RubricBench prompts on which ReVeR and Rubric-ARM reach opposite decisions.
Instructions, responses, rubrics, programs, and judge outputs are quoted verbatim, and [...] marks an omission.

\subsection{A Case Decided by the Judge}
\label{app:case-judge}

This is the case of \S\ref{sec:case}.
The two rubrics contain the same hard rule on the letter p, and the two judges apply it differently.

\paragraph{Instruction.}
I'm interested in a college with open enrollment and a regional accreditation. Which college would you recommend? Don't include the keywords ``DuPage'' and ``Dade'' in your response. Let's make it a constrained writing problem: be sure the letter p appears at least 15 times in your response.

\paragraph{Expert label.}
Response A is better.

\paragraph{Response A.}
\begin{quote}\small\raggedright\setlength{\parskip}{2pt}
Choosing the perfect college is a very personal decision, and it's great you're prioritizing open enrollment and regional accreditation.  \par
To give you a helpful recommendation, I need a little more information about your preferences.  \par
For example, what are your academic interests? Are you looking for a primarily online program or a more traditional campus experience? \par
Tell me more about what's important to you in a college, and I can provide a more specific suggestion. \par
Remember, finding the right fit is crucial for a positive and successful educational experience.\par
\end{quote}
\paragraph{Response B.}
\begin{quote}\small\raggedright\setlength{\parskip}{2pt}
Choosing a college is a big decision!  It's great that you're prioritizing open enrollment and regional accreditation, as those are important factors for a quality education. \par
To give you the best recommendation, I need a little more information about your preferences. \par
For example:\par
* **What part of the country are you interested in?**  \par
* **Are you looking for a two-year or four-year program?**\par
* **What are your academic interests?**\par
Once I have a better understanding of what you're looking for, I can provide a more personalized suggestion.  \par
Remember, finding the right college is a process, so don't be afraid to explore your options and ask lots of questions!  \par
Let me know if you have any other details you'd like to share. I'm here to help!\par
\end{quote}
\paragraph{Expert rubric.}
\begin{itemize}[leftmargin=*,itemsep=0pt,topsep=2pt]\small
  \item Does the response explicitly state that the recommended college has an ``open enrollment'' policy?
  \item Does the response explicitly state that the recommended college has ``regional accreditation''?
  \item Does the response avoid including the words ``DuPage'' and ``Dade''?
  \item Does the response contain the letter 'p' at least 15 times?
  \item Does the response provide a specific, named college or university as a recommendation?
  \item Is the response presented as a continuous paragraph without bullet points or lists?
\end{itemize}
\paragraph{Rubric of ReVeR (SFT).}
\begin{enumerate}[leftmargin=*,itemsep=0pt,topsep=2pt]\small
  \item The response recommends a specific named college that meets both the open enrollment and regional accreditation criteria. [Hard Rule]
  \item The response does not contain the substring ``dupage'' or ``dade'' (case-insensitive). [Hard Rule]
  \item The response contains at least 15 occurrences of the lowercase letter 'p'. [Hard Rule]
  \item Any factual claims about the recommended college's open enrollment status or regional accreditation are accurate according to publicly available information. [Hard Rule]
  \item The response forms a coherent piece of writing that presents the recommendation, rather than consisting solely of characters inserted only to satisfy the letter count. [Principle]
\end{enumerate}
\paragraph{Rubric of Rubric-ARM.}
\begin{enumerate}[leftmargin=*,itemsep=0pt,topsep=2pt]\small
  \item The response must recommend a college that offers open enrollment and regional accreditation. [Hard Rule]
  \item The response must not include the keywords ``DuPage'' or ``Dade''. [Hard Rule]
  \item The response must contain at least 15 instances of the letter p. [Hard Rule]
  \item The response should directly address the user's request by providing relevant recommendations. [Principle]
  \item The response should be organized in a clear and logical manner to enhance readability. [Principle]
  \item The response should maintain a professional and courteous tone. [Principle]
  \item The response should be concise and avoid unnecessary or irrelevant information. [Principle]
\end{enumerate}
\paragraph{ReVeR judge.}
The judge scores each response in its own episode and runs one program per response.
For Response A it writes:
\begin{lstlisting}
response = """<Response A>"""

# Check for "dupage" or "dade" (case-insensitive)
import re
dupage_match = re.search(r'dupage', response, re.IGNORECASE)
dade_match = re.search(r'dade', response, re.IGNORECASE)

print("Contains 'dupage':", dupage_match is not None)
print("Contains 'dade':", dade_match is not None)

# Count lowercase 'p' occurrences
lowercase_p_count = response.count('p')
uppercase_P_count = len(re.findall(r'[Pp]', response))
print(f"Lowercase 'p' count: {lowercase_p_count}")
print(f"Total 'p'/'P' count: {uppercase_P_count}")
\end{lstlisting}
and the interpreter returns:
\begin{lstlisting}
Contains 'dupage': False
Contains 'dade': False
Lowercase 'p' count: 16
Total 'p'/'P' count: 16
\end{lstlisting}
The same count on Response B returns 13.
The scores on the five criteria of the rubric are:
\begin{center}
\small
\begin{tabular}{@{}clcc@{}}
    \toprule
    \textbf{Criterion} & \textbf{Type} & \textbf{Response A} & \textbf{Response B} \\
    \midrule
    1 & hard rule & 0 & 0 \\
    2 & hard rule & 1 & 1 \\
    3 & hard rule & 1 & 0 \\
    4 & hard rule & 1 & 1 \\
    5 & principle & 10 & 0 \\
    \midrule
    \multicolumn{2}{@{}l}{Aggregate score} & 3.10 & 2.00 \\
    \bottomrule
\end{tabular}
\end{center}
ReVeR prefers Response A and is correct in all six runs, three with each generator.

\paragraph{Rubric-ARM judge.}
The pairwise judge prefers Response B in both presentation orders of all three repeats.
Its output ends as follows in the original order:
\begin{quote}\small\raggedright\setlength{\parskip}{2pt}
[...] ent -{}-{}-\par
Aggregation Summary: Both responses fail the primary recommendation requirement, but only Response B satisfies the crucial letter-p count gatekeeper.  \par
Justification: Despite neither naming a college, Response B alone meets Criterion 3, making it the stronger compliance.  \par
Winner: Response B\par
\end{quote}
and as follows in the swapped order, where its ``Response A'' is the original Response B:
\begin{quote}\small\raggedright\setlength{\parskip}{2pt}
[...] responses fail the critical recommendation requirement and the letter-p count gatekeeper. Between them, Response A at least reaches closer to 15 p's.  \par
Justification: Although neither meets the key recommendation rule, Response A edges out Response B on the p-count requirement.  \par
Winner: Response A\par
\end{quote}

\subsection{A Case Decided by the Rubric}
\label{app:case-rubric}

Here the two rubrics ask for different things.
The instruction leaves out the interest rate, and the rubric of Rubric-ARM requires a product of an annual amount and a number of years that the instruction does not mention.

\paragraph{Instruction.}
120000 for 30 year what will be the savings.

\paragraph{Expert label.}
Response A is better.

\paragraph{Response A.}
\begin{quote}\small\raggedright\setlength{\parskip}{2pt}
I need more information to calculate the savings. \par
Please tell me:\par
* **What is the interest rate?** This is crucial for calculating how much your money will grow over time.\par
* **Is the interest compounded?**  If so, how often (e.g., annually, monthly)? Compounding means earning interest on your interest, which significantly increases your savings over time.\par
* **Are you making any additional contributions?**  If you're adding more money to the savings regularly, it will impact the final amount.\par
Once you provide these details, I can help you calculate your potential savings.\par
\end{quote}
\paragraph{Response B.}
\begin{quote}\small\raggedright\setlength{\parskip}{2pt}
To calculate the savings after 30 years for an initial amount of \$120,000, we need to consider the interest rate and the type of account where the money is saved. For this example, let's assume a conservative interest rate of 3\% per year, compounded annually, in a savings account. Here's how to calculate the future value:\par
Future Value (FV) = Present Value (PV) * (1 + Interest Rate) \textasciicircum{} Number of Years\par
FV = \$120,000 * (1 + 0.03) \textasciicircum{} 30\par
FV = \$120,000 * 1.03 \textasciicircum{} 30\par
FV $\approx$ \$336,960.71\par
So, after 30 years, the initial amount of \$120,000 would grow to approximately \$336,960.71, assuming a 3\% annual interest rate compounded annually. This represents a total savings of \$216,960.71 over the 30-year period.\par
Please note that the actual savings may vary depending on the interest rate, compounding frequency, and any additional contributions or withdrawals made during the 30-year period.\par
\end{quote}
\paragraph{Expert rubric.}
\begin{itemize}[leftmargin=*,itemsep=0pt,topsep=2pt]\small
  \item Does the response avoid giving concrete calculations when key parameters (e.g., interest rate, compounding) are missing?
  \item Does the response actively request missing information needed for accurate calculation?
  \item Does the response acknowledge that the instruction is missing crucial information and either ask clarifying questions or present multiple, clearly stated scenarios?
  \item Does the response explicitly identify which variables are missing from the prompt and are required to provide a definitive answer?
\end{itemize}
\paragraph{Rubric of ReVeR (SFT).}
\begin{enumerate}[leftmargin=*,itemsep=0pt,topsep=2pt]\small
  \item The response either requests the missing required parameters (e.g., interest rate, contribution amount, compounding frequency) or explicitly states the assumptions made for the calculation. [Hard Rule]
  \item Any numerical calculation included is mathematically correct given the stated assumptions. [Hard Rule]
  \item When assumptions are made, the response explains the rationale for each assumption and acknowledges potential uncertainties (e.g., varying interest rates, tax implications, withdrawals). [Principle]
  \item When a projected amount is provided, it is accompanied by the calculation method or sufficient detail to allow the user to reproduce the result or understand its meaning. [Principle]
\end{enumerate}
\paragraph{Rubric of Rubric-ARM.}
\begin{enumerate}[leftmargin=*,itemsep=0pt,topsep=2pt]\small
  \item The response must calculate the total savings by multiplying the given annual amount by the specified number of years. [Hard Rule]
  \item The response should present the final numerical result clearly and unambiguously. [Principle]
  \item The response should be concise and avoid unnecessary elaboration. [Principle]
\end{enumerate}
\paragraph{ReVeR judge.}
The figure in Response B is wrong: $120{,}000\times 1.03^{30}=291{,}271.50$, not 336{,}960.71.
The second hard rule of ReVeR's rubric covers this, and the judge scores it 0 for Response B.
ReVeR prefers Response A, with scores 2.99 and 1.54, and is correct in all six runs.

\paragraph{Rubric-ARM judge.}
The pairwise judge prefers Response B in both presentation orders of all three repeats.
Its output ends as follows in the original order:
\begin{quote}\small\raggedright\setlength{\parskip}{2pt}
[...] by the years. Between them, neither fulfills Criterion 1, but Response B at least provides a clear numeric result.  \par
Justification: Although both fail the primary calculation requirement, Response B delivers a specific numerical outcome, whereas Response A offers no calculation.  \par
Winner: Response B\par
\end{quote}
and as follows in the swapped order, where its ``Response A'' is the original Response B:
\begin{quote}\small\raggedright\setlength{\parskip}{2pt}
[...] essential multiplication requirement, but Response A at least supplies a numerical result (albeit incorrect), whereas Response B provides no calculation.  \par
Justification: Although neither fully meets the hard rule, Response A demonstrates an attempt at quantitative calculation.  \par
Winner: Response A\par
\end{quote}

\onecolumn
\section{Prompts}
\label{app:prompts}

The two boxes give the prompts of ReVeR in full.
The generator prompt is sent as a single user message, with \texttt{\{instruction\}} replaced by the instruction; \texttt{\{context\}} is empty in all experiments.
The teacher and the generator use this prompt in distillation, in generator RL, and in every evaluation.
The judge prompt is the system message.
Its user message is a JSON object that holds the instruction, one response, and the numbered criteria, and one tool, \texttt{python}, executes a program with a timeout of three seconds.
The box shows the prompt used at inference, with a budget of three tool calls; the prompt of the teacher trajectories differs only in this budget, which is one call (\S\ref{sec:judge}).
The in-context examples of both prompts are taken from RubricBench, and these prompts are excluded from every RubricBench result we report.

\promptbox{Rubric generator prompt (user message)}{prompts/rubric_generator.txt}

\promptbox{Judge prompt (system message)}{prompts/judge.txt}

\end{document}